\documentclass[letterpaper, 10 pt, conference]{ieeeconf}  

\IEEEoverridecommandlockouts                              

\title{\LARGE \bf
Learning to Identify Object Instances by Touch:\\ Tactile Recognition via Multimodal Matching
}

\author{Justin~Lin$^{1}$, Roberto~Calandra$^{2}$, and Sergey~Levine$^{1}$%
\thanks{*This work was supported by Berkeley DeepDrive and the Office of Naval Research (ONR).}%
\thanks{$^{1}$Department of Electrical Engineering and Computer Sciences, University of California, Berkeley, USA\newline
 {\tt\small justinlin98@berkeley.edu,\newline  svlevine@eecs.berkeley.edu}}%
\thanks{$^{2}$Facebook AI Research, Menlo Park, CA, USA\newline
        {\tt\small rcalandra@fb.com}}%
}

\usepackage[english]{babel}

\usepackage{graphicx}
\usepackage{animate}
\usepackage{epsfig} 				
\usepackage{epstopdf}

\usepackage{listings}
\usepackage{color}
\usepackage{nameref}
\usepackage{hyperref}
\usepackage[colorinlistoftodos]{todonotes}
\usepackage{amsmath}	 			
\usepackage{amssymb}  				

\usepackage{dsfont}			
\usepackage{mathtools}

\usepackage{epigraph}
\usepackage{lscape}
\usepackage[]{nomencl}				
\usepackage[ruled,vlined,linesnumbered]{algorithm2e}
\usepackage{multicol}
\usepackage{multirow}
\usepackage{etoolbox}
\usepackage{siunitx}

\usepackage{caption}
\usepackage{subcaption}

\usepackage{wrapfig}
\usepackage{multimedia}

\usepackage{units}
\usepackage{flushend}

\usepackage{floatflt}

\usepackage{url}

\usepackage[absolute,overlay]{textpos}

\usepackage{bibentry}

\usepackage{tikz}
\usepgflibrary{arrows}	

\usepackage{float}
\usepackage[utf8]{inputenc}
\usepackage[english]{babel}
\usepackage{graphics}
\usepackage{animate}

\usepackage{listings}

\usepackage[nostamp]{draftwatermark}

\usepackage{extarrows}		

\allowdisplaybreaks[1]

\newcommand{\playvideo}[1]{\href{run:#1}{\includegraphics[scale=0.12]{\RCPath fig/empty}}}

\usepackage{lipsum}
\usepackage{framed}

\newcommand{\expec}[1]{\mathbb{E}_{{#1}}}

\newcommand{\email}[1]{\href{mailto:#1}{\nolinkurl{#1}}}

\newcommand{\link}[1]{\colora{\url{#1}}}

\newcommand{\fig}[1]{Figure~\ref{#1}}

\newcommand{\tab}[1]{Table~\ref{#1}}

\definecolor{matlab1}{rgb}{0,0,1}
\definecolor{matlab2}{rgb}{0,0.5,0}
\definecolor{matlab3}{rgb}{1,0,0}
\definecolor{matlab4}{rgb}{0,0.75,0.75}
\definecolor{matlab5}{rgb}{0.75,0,0.75}
\definecolor{matlab6}{rgb}{0.75,0.75,0}
\definecolor{matlab7}{rgb}{0.25,0.25,0.25}

\definecolor{darkgreen}{rgb}{0,0.5,0}		
\definecolor{purple}{rgb}{0.75,0,0.75}
\definecolor{pink}{rgb}{1,0.4,0.6}

\newcommand{\capitalize}[1]{\expandafter\MakeUppercase\expandafter{#1}}

\newcommand{\colora}[1]{{\usebeamercolor[fg]{framesubtitle}#1}}

\makeatletter
\newcommand*{\compress}{\@minipagetrue}
\makeatother

\renewcommand{\vec}[1]{\boldsymbol{#1}}				

\newcommand{\q}{\vec{q}}					
\ifdef{\dq}{\renewcommand{\dq}{\dot{\q}}}{\newcommand{\dq}{\dot{\q}}}

\usepackage{array}
\makeatletter
\newcommand{\thickhline}{%
    \noalign {\ifnum 0=`}\fi \hrule height 1pt
    \futurelet \reserved@a \@xhline
}
\newcolumntype{"}{@{\hskip\tabcolsep\vrule width 1pt\hskip\tabcolsep}}
\makeatother

\newcommand{\citet}[1]{\cite{#1}}
\newcommand{\citep}[1]{\cite{#1}}

\begin{document}
\maketitle


\begin{abstract}     
Much of the literature on robotic perception focuses on the visual modality. 
Vision provides a global observation of a scene, making it broadly useful. 
However, in the domain of robotic manipulation, vision alone can sometimes prove inadequate: in the presence of occlusions or poor lighting, visual object identification might be difficult. 
The sense of touch can provide robots with an alternative mechanism for recognizing objects. 
In this paper, we study the problem of touch-based instance recognition. 
We propose a novel framing of the problem as multi-modal recognition: the goal of our system is to recognize, given a visual and tactile observation, whether or not these observations correspond to the same object. 
To our knowledge, our work is the first to address this type of multi-modal instance recognition problem on such a large-scale with our analysis spanning 98 different objects. 
We employ a robot equipped with two GelSight touch sensors, one on each finger, and a self-supervised, autonomous data collection procedure to collect a dataset of tactile observations and images. 
Our experimental results show that it is possible to accurately recognize object instances by touch alone, including instances of novel objects that were never seen during training. Our learned model outperforms other methods on this complex task, including that of human volunteers.

\end{abstract}


\section{Introduction}

	Imagine rummaging in a drawer, searching for a pair of scissors. You feel a cold metallic surface, but it's smooth and curved -- that's not it. You feel a curved plastic handle -- maybe? A straight metal edge -- you've found them! Humans naturally associate the appearance and material properties of objects across multiple modalities. Our perception is inherently \emph{multi-modal}: when we see a soft toy, we imagine what our fingers would feel touching the soft surface, when we feel the edge of the scissors, we can picture them in our mind -- not just their identity, but also their shape, rough size, and proportions. Indeed, the association between visual and tactile sensing forms a core part of our manipulation strategy, and we often prefer to identify objects by touch rather than sight, either when they are obscured, when our gaze is turned or elsewhere, or simply out of convenience. In this work, we study how similar multi-modal associations can be learned by a robotic manipulator. We frame this problem as one of cross-modality instance recognition: recognizing that a tactile observation and a visual observation correspond to the same object instance.
\begin{figure}
	\centering
	\includegraphics[width=0.99\linewidth]{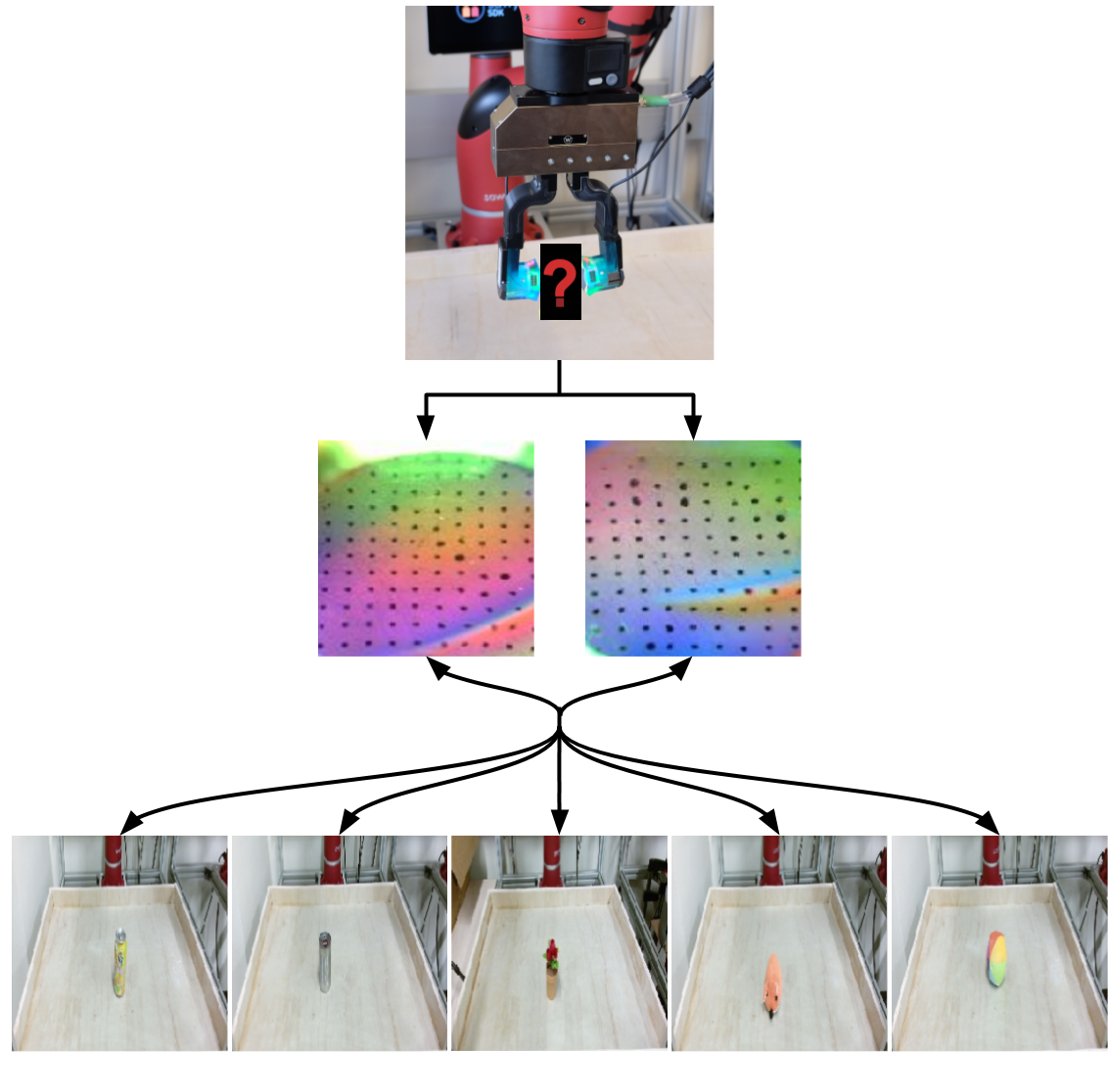}
	\caption{Illustration of the cross-modality object recognition problem. 
	When grasping an unknown object, given a pair of GelSight touch sensors (top) and candidate object images (bottom), the robot must determine which object the tactile readings correspond to.}
	\label{fig:setting}
	\vspace{-10pt}
\end{figure}
This type of cross-modal recognition has considerable practical value. By enabling a robot to identify objects by touch, robots can pick up and manipulate objects even when visual sensing is obscured. For example, a warehouse automation robot might be able to retrieve a particular object from a shelf by feeling for it with its fingers, matching the tactile observations to a product image from the manufacturer. We might also expect tactile recognition to generalize better than visual recognition, since it suffers less from visual distractors, clutter, and illumination changes. However, this problem setting also introduces some very severe challenges. First, tactile sensors do not have the same kind of global view of the scene as the visual modality, which means that the cross-modal association must be made by matching very local properties of a surface to an object's overall appearance. Second, tactile readings are difficult to interpret.
In \fig{fig:setting}, we have photographs of five different objects, and readings from the fingers of a parallel jaw gripper equipped with GelSight touch sensors~\cite{Yuan2014}. 
Can you guess which object is being grasped?
We show below that even for humans, it might not be obvious from tactile readings which object is being touched.

Our approach to the cross-modality instance recognition problem is based on high-resolution touch sensing using the GelSight sensor~\cite{Yuan2014} and the use of convolutional neural network models for multi-modal association. The GelSight sensor produces readings by means of a camera embedded in an elastomer gel, which observes indentations in the gel made by contact with objects. Since the readings from this sensor are represented as camera images, it is straightforward to input them into standard convolutional neural network models, which have proven extremely proficient at processing visual data. We train a convolutional network to take in the tactile readings from two GelSight sensors which are mounted on the fingers of a parallel jaw gripper, as well as an image of an object from a camera, and predict whether these inputs come from the same object or not. By combining the visual observation of the query object with the robot's current tactile readings we are able to perform instance recognition. 

However, as with all deep learning based methods, this approach requires a large dataset for training. A major advantage of our approach is that, unlike supervised recognition of object categories, the cross-modality instance recognition task can be self-supervised with autonomous data collection. Since semantic class labels are never used during training, the robot can collect a large dataset by grasping objects in the environment and recording the image before the grasp and the tactile observations during the grasp. Repeated grasps of the same object can be used to build a dataset of positive examples (by associating every tactile reading for an object with every image for that object), while all other pairings of images and touch readings in the dataset can be used as negative examples. 

Our main contributions are to formulate the cross-modality instance recognition problem, propose a solution based on deep convolutional neural networks along with an autonomous data collection procedure that requires minimal human supervision, and provide empirical results on a large-scale dataset with over 90 different objects which demonstrate cross-modality instance recognition with vision and touch is indeed feasible.


\section{Related Work}

	To our knowledge, our work is the first to propose the cross-modality instance recognition problem with vision and touch for \emph{object} recognition in particular. However, a number of prior works have studied related tactile recognition problems. The most closely related is the work of Yuan et al. \cite{Yuan2017a}, which trains a model to detect different types of cloth using touch, vision, and depth. Our model is closely related to the one proposed by Yuan et al. \cite{Yuan2017a}, but the problem setting is different: the goal in this prior work is to recognize types of cloth, where local material properties are broadly indicative of the object identity (indeed, cloth type in this work is \emph{determined} by material properties). In contrast, we aim to recognize object instances. The fact that this is even possible is not at all obvious. While we might expect touch sensing to help recognize materials, object identity depends on global geometry and appearance properties, and it is not obvious that touch provides significant information about this. Our experiments show that in fact it does.

A number of prior works have also explored other multi-modal association problems, such as audio-visual matching~\cite{nagrani2018seeing}, visual-language matching~\cite{kiros2014multimodal} and three-way trajectory-audio-visual~\cite{droniou2015deep}. Our technical approach is similar, but we consider self-supervised association of vision and touch. We use a neural network model inspired by that of \cite{arandjelovic2017look}, which trained a two-stream classifier to predict whether images and audio come from the same video.

Touch sensing has been employed in a number of different contexts in robotic perception. Kroemer et al. \cite{Kroemer2011} proposed matching time series of tactile measurements to surface textures using kernel machines. In contrast, our method recognizes object instances, and uses an observation space that has much higher dimensionality (by about two orders of magnitude). More recently, researchers have also proposed other, more strongly supervised techniques for inferring object properties from touch. For example~\cite{Yuan2017} proposed estimating the hardness of an object using a convolutional network, while~\cite{Li2013,murali2018learning,luo2018vitac} estimated material and textural properties. However, to our knowledge, none of these prior works have demonstrated that object instances (rather than just material properties) can be recognized entirely by touch and matched to corresponding visual observations.

Aside from recognition and perception, tactile information has also been extensively utilized for directly performing robotic manipulation skills, especially grasping. For example, \cite{Calandra2017,Calandra2018,Hogan2018,murali2018learning} predicted how suitable a given gripper configuration was for grasping. We take inspiration from these approaches, and use the tactile exploration strategy of \cite{Calandra2017}, whereby the robot ``feels'' a random location of an object using a two-fingered gripper equipped with two GelSight~\cite{Johnson2011,Dong2017} touch sensors.

The association between vision and touch has also been considered in psychology. One of the earliest examples of this is the Molyneux problem~\cite{locke1841essay}, which asks whether a blind person---upon gaining the ability to see---could match geometric shapes to their tactile stimuli. Later experimental work~\cite{held2011newly} has confirmed that the association between touch and sight in fact requires extensive experience with both modalities.


\section{Associating Sight and Touch}

	The goal of our approach is to determine, given a visual observation and a tactile observation, whether or not these two observations correspond to the same object. A model that can answer this question can then be used to recognize individual objects: given images of candidate objects, the robot can test the association between its current tactile observations and each of these object images, predicting the object with the highest probability of correspondence as the object currently being grasped.
It is worth noting that with our setup, all images come from the same environment. Generalizing this system to multiple environments would likely require a more diverse data-collection effort or domain adaptation methods~\cite{Hoffman2016}, although we expect the basic principles to remain the same.

\subsection{Task Setup and Data Collection}
\begin{wrapfigure}{r}{0.52\linewidth}
	\centering
	\vspace{-10pt}
	\includegraphics[width=0.98\linewidth]{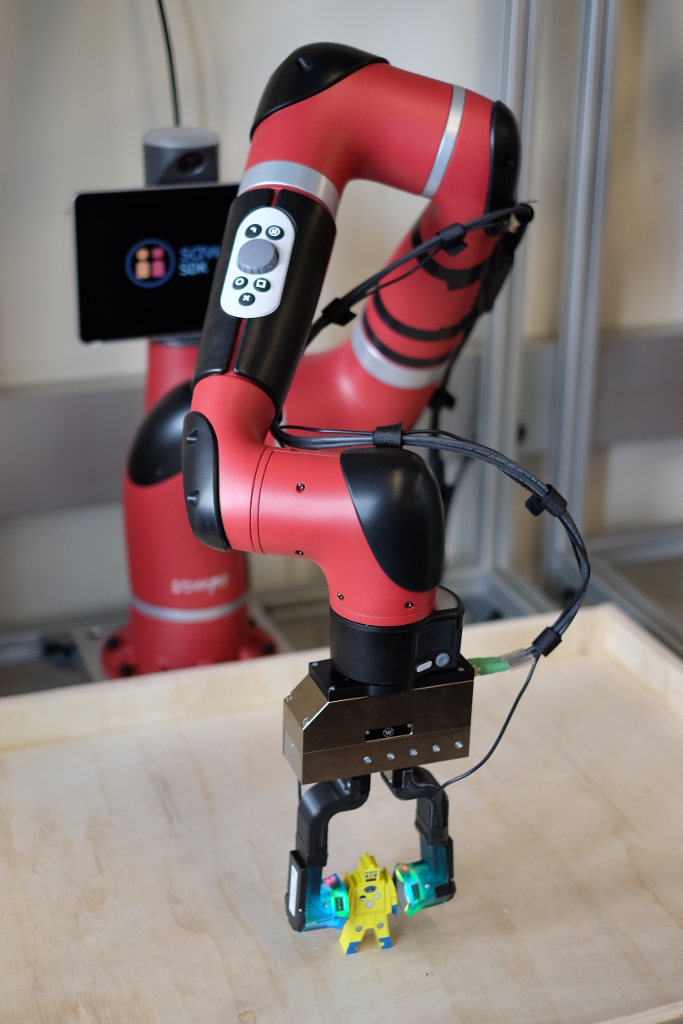}
	\caption{Our experimental setting consists of two GelSight tactile sensors mounted on a parallel jaw gripper, and a frontal RGB camera.}
	\label{fig:robot}
\end{wrapfigure}
Though the basic setup of our method could in principle be applied to any tactile sensor, we hypothesize that high-resolution surface sensing is important for successfully recognizing objects by touch. We therefore utilize the GelSight~\cite{Yuan2017b} sensor, which consists of a deformable gel mounted above a camera. The gel is illuminated on different sides, such that the camera can detect deformations on the underside of the gel caused by contact with objects. This type of sensor produces high-resolution image observations, and can detect fine surface features and material details. In our experimental setup, shown in \fig{fig:robot}, two GelSight sensors are mounted on the fingers of a parallel jaw gripper, which interacts with objects by attempting to grasp them with top-down pinch grasps. The images from the sensors are downsampled to a resolution of $128 \times 128$ pixels. 
Since the GelSight sensor produces ordinary RGB images, we can process these readings with conventional convolutional neural network models.

\begin{figure}[t]
  \centering
		\includegraphics[width=0.32\linewidth]{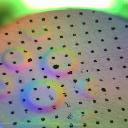}
		\includegraphics[width=0.32\linewidth]{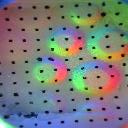}
		\includegraphics[width=0.32\linewidth]{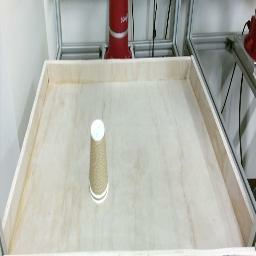}
        \\\vspace{5pt}
        \includegraphics[width=0.32\linewidth]{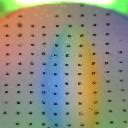}
		\includegraphics[width=0.32\linewidth]{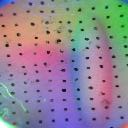}
		\includegraphics[width=0.32\linewidth]{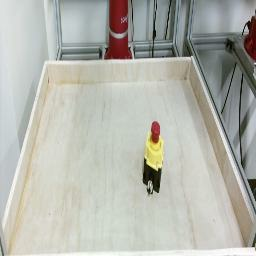}
        \\\vspace{5pt}
        \includegraphics[width=0.32\linewidth]{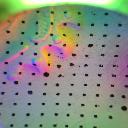}
		\includegraphics[width=0.32\linewidth]{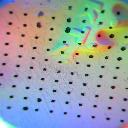}
		\includegraphics[width=0.32\linewidth]{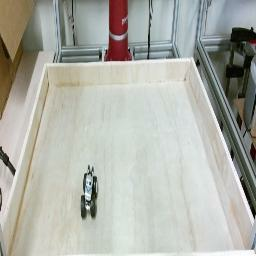}
  \caption{Examples of tactile readings and object images which correspond to a single object. These images form the $(T, I)$ pairs which are fed into our network.}
  \label{fig:example}
\end{figure}

The visual observation is recorded by a conventional RGB camera -- in our case, the camera on a Kinect 2 sensor (note that while we utilize the depth sensor for data collection we do not use the depth observation at all for recognition). This camera records a frontal view of the object, as shown in \fig{fig:example}.

The cross-modality instance recognition task therefore is defined as determining whether a given tactile observation~$T$, represented by two concatenated $128 \times 128$ 3-channel images from the GelSight sensors, corresponds to the same object as a visual observation~$I$, represented by a $256 \times 256$ RGB image from the Kinect.

We collect data for our method using a 7-DoF Sawyer arm, to which we mount a Weiss WSG-50 parallel jaw gripper equipped with the GelSight sensors. For collecting the data, we use an autonomous and self-supervised data collection setup that allows the robot collect the data unattended, only requiring human intervention to replace the object on the table.\footnote{Human intervention can be removed entirely if for example the robot autonomously grasps objects from a bin.} At the beginning of each interaction, we record the visual observation $I$ from the camera -- while we considered cropping these images, we find that leaving the images unmodified results in better performance. We then use the depth readings from the Kinect 2 to fit a cylinder to the object and move the end-effector to a random position centered at the midpoint of this cylinder. Next we close the fingers with a uniformly sampled force and record the tactile readings $T$. We only consider tactile readings for which the grasp is successful; this is determined by a deep neural network classifier similar to~\cite{Calandra2017} that is trained to recognize tactile readings $T$ which come from successful grasps. Data is collected for 98 different objects, and these objects are randomly divided into a training set of 80 objects and a test set of 18 objects respectively. We take care to use an equal amount of examples from each object when training and evaluating the model. Positive examples for training are constructed from random pairs $T_i, I_j$, where $i$ and $j$ index interactions with the same object. Negative examples are constructed from random pairs $T_i, I_j$ where $i$ indexes one object, and $j$ indexes another object in the same set. We balance the dataset such that there are an equal number of positive and negative examples.

\subsection{Convolutional Networks for Cross-Modality Instance Recognition}
\begin{figure*}[t]
  \centering
  \includegraphics[width=\linewidth]{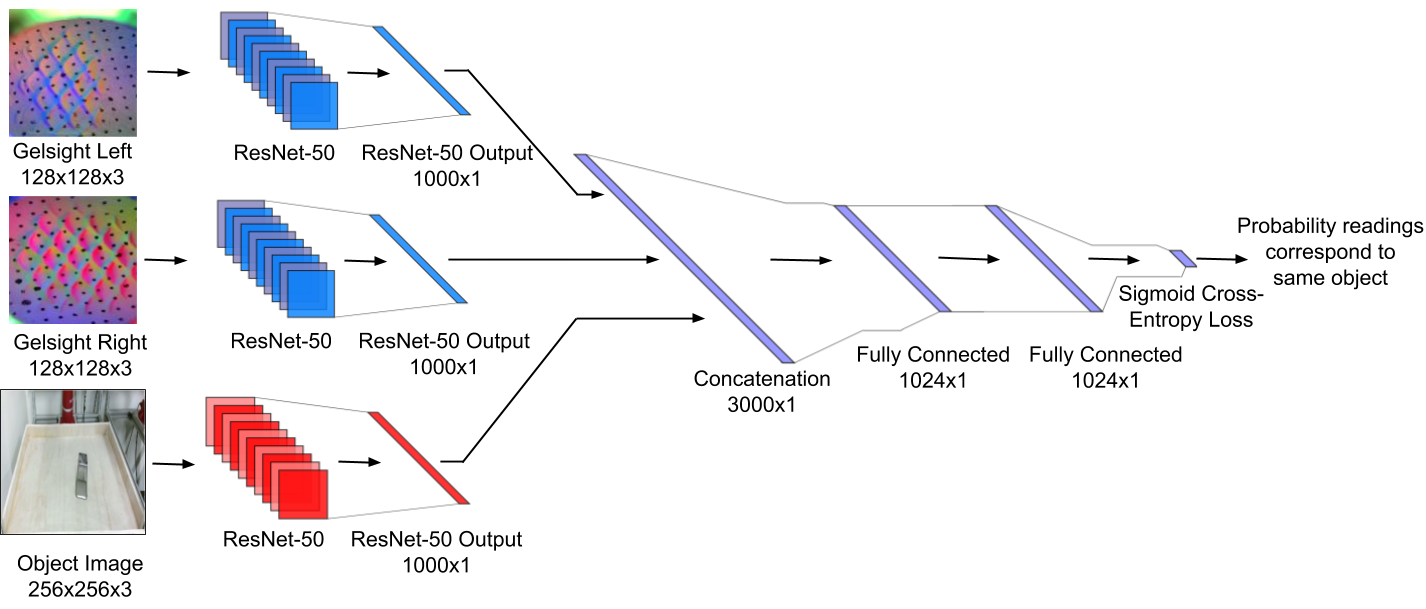}
  \caption{High-level diagram of our cross-modality instance recognition model. ResNet-50 CNN blocks are used to encode both of the tactile readings and the visual observation. Note that the weights of the ResNet-50s for the two tactile readings are tied together. The features from all modalities are fused via concatenation and passed through 2 fully connected layers before outputting the probability that the readings match.}
    \label{fig:model}
\end{figure*}
Our model is trained to predict whether a tactile input $T$ and a visual image $I$ correspond to the same object. Given the dataset described in the previous section, this can be accomplished by using a maximum likelihood objective to optimize the parameters $\theta$ for a model of the form $p_\theta(y | T, I)$, where $y$ is Bernoulli random variable that indicates whether $T$ and $I$ map to the same object. The objective is given by
\begin{eqnarray}
\mathcal{L}(\theta) &=& \expec{(T_i, I_j) \in \mathcal{D}_{\footnotesize \mbox{same}}}[\log(p_\theta(y = 1 \mid T_i, I_j))]  \\ 
&& ~+~\expec{(T_i, I_j) \in \mathcal{D}_{\footnotesize \mbox{diff}}}[\log(p_\theta(y = 0 \mid T_i, I_j))]\,, \nonumber
\end{eqnarray}
where $\mathcal{D}_{\footnotesize \mbox{same}}$ and $\mathcal{D}_{\footnotesize \mbox{diff}}$ are the sets of visuo-tactile examples that come from the same or different objects respectively.

The particular convolutional neural network that we use to represent $p_\theta(y | T, I)$ in our method is illustrated at a high level in Figure~\ref{fig:model}. Since all of the inputs are represented as images, we first encode all of the images using a ResNet-50 convolutional network backbone~\cite{He2016}. We employ a late fusion architecture, where both of the GelSight images and the visual observation are fused after the convolutional network by concatenating the final (after the last fully connected layer) outputs of all three ResNet-50 backbones giving us a joint visuo-tactile feature representation of 3000 units total, which is then passed through 2 more fully connected layers. Each of these fully connected layers has 1024 hidden units with ReLU nonlinearities and we perform dropout regularization between the two layers. After the last fully connected layer, the network outputs a class probability via a sigmoid for the positive and negative class, which indicates whether or not $T$ and $I$ correspond to the same object. Since both of the GelSight images represent the same modality, we tie the weights of the ResNet-50 blocks that featurize the two tactile readings.

As mentioned above, for training we feed in pairs ($T$, $I$) in which each tactile input is paired with eight random visual inputs: four positive examples from the same object and four
negative examples from different objects in the training set. We train the model using the Adam optimizer~\cite{kingma2014adam} with an initial learning rate of $10^{-4}$ for 26,000 iterations and a batch size of 48. The ResNet-50 blocks for both the tactile and visual branches of the network are pretrained on the ImageNet object recognition task~\cite{deng2009imagenet} to improve invariance and speed up convergence.

\subsection{Recognizing Object Instances}
\label{sec:kway}

Our cross-modality instance recognition model can be used in several different ways. One such application is to simply evaluate how confident our model is that a given object image corresponds to a given tactile reading. However, in practice, we might like to use this model to recognize object instances by touch. We can use the model for this without additional training, as follows. First, we need to obtain a set of candidate object images. In a practical application, these candidate images might come from product images from a manufacturer or retailer, but in our case the images come from a test set of grasps recorded in the same environment. Then we can select which object the robot is most likely grasping by predicting
\begin{align}
k^\star = \arg\max_k \quad \log p_\theta(y = 1 | T, I_k)\,.
\end{align}
In our experiments, we perform object identification in this manner.


\section{Experimental Results}

We aim to understand whether our method can recognize specific object instances for unseen test objects through our experimental evaluation. Note that this task is exceptionally challenging: in contrast to material and surface recognition, which has been explored as a potential application of touch sensing, object recognition potentially requires non-local information about shape and appearance that is difficult to obtain from individual tactile readings.

\subsection{Matching Vision and Touch}

\begin{figure}[t]
  \centering
  \includegraphics[height=4.5cm]{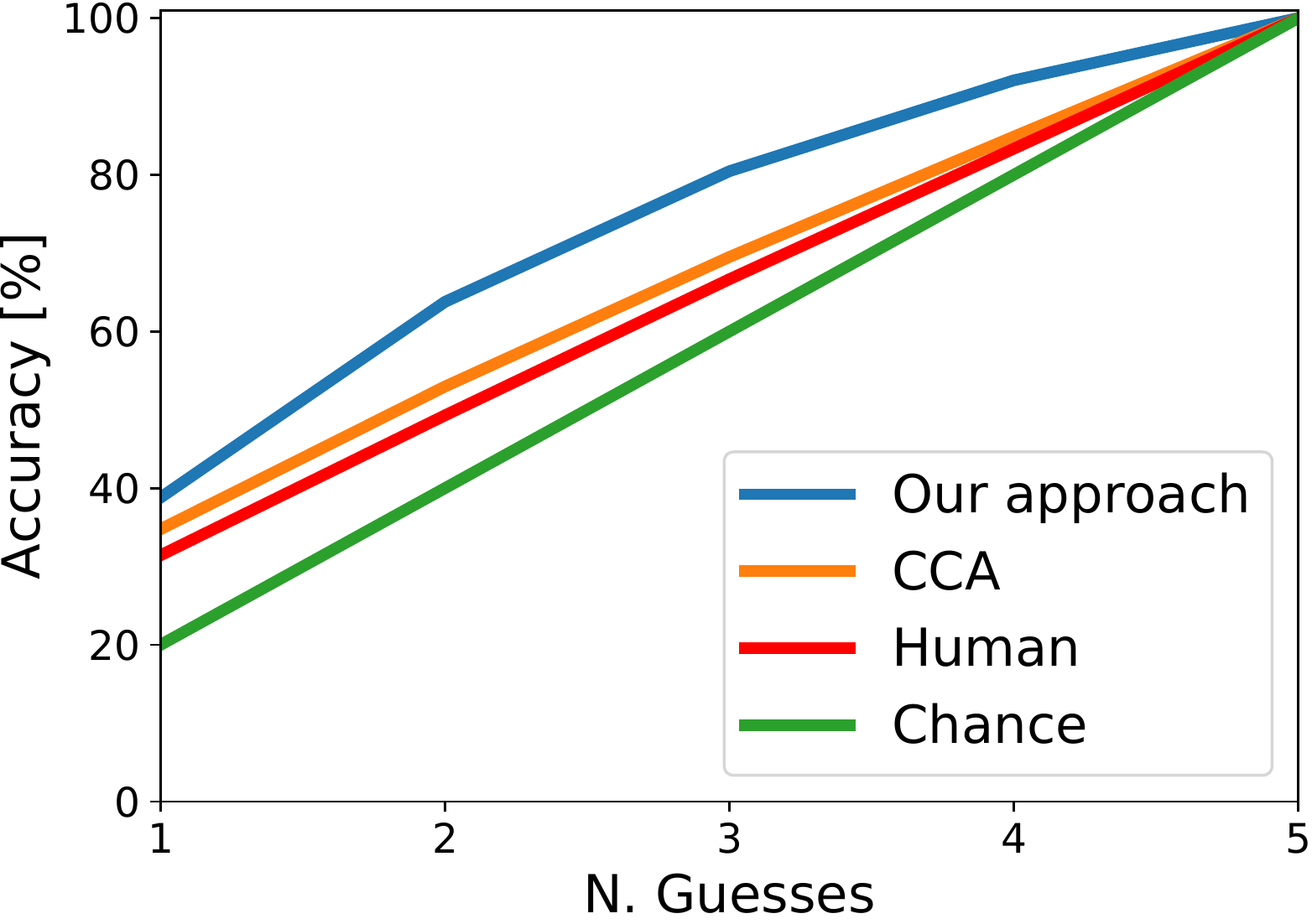}
  \caption{Accuracy for 5-shot classification. Our model outperforms both CCA and human performance. For both CCA and the humans, after a strong first guess the increase in accuracy beyond chance is fairly modest. On the other hand, our cross-modality instance recognition model continues to achieve recognizable gains beyond the first guess, suggesting that it has learned a meaningful association between vision and touch for many of the objects.}
    \label{fig:roc5}
\end{figure}

\begin{table}[b]
			\centering
			\label{tab:model}
	    \begin{tabular}{|c|c|c|}
			  \hline 
              Method & Accuracy\\
              \hline
              Our model & 64.3\%\\
              Chance & 50.0\%\\
			  \hline 
			\end{tabular}\\
			\caption{Model accuracy for direct object-instance recognition. Given a pair ($T, I$) of tactile readings and object image our model predicts whether both elements in the pair correspond to the same object.}
    \label{tab:accuracy}
\end{table}

\begin{figure}[t]
  \centering
  \includegraphics[height=4.5cm]{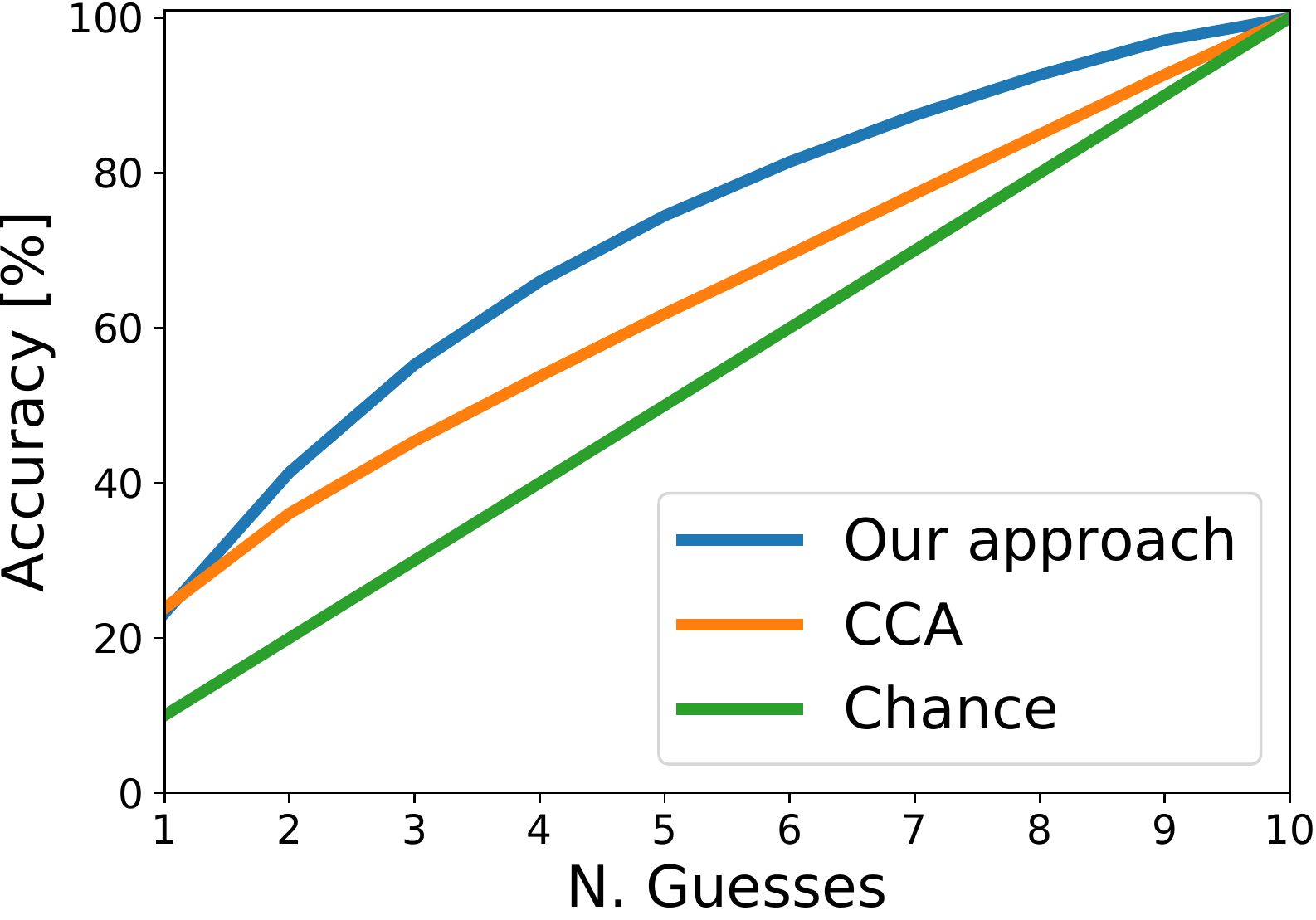}
  \caption{Accuracy for 10-shot classification. The accuracy curves are quite similar to the $K = 5$ case in which CCA makes a good initial guess but afterwards fares no better than random chance. Meanwhile our model is able to make intelligent predictions even when it is not correct on its first attempt, which explains its higher accuracy numbers when compared to the benchmark.}
    \label{fig:roc10}
\end{figure}

We first analyze the performance of our model directly. Using the data generation procedure mentioned above, we obtain a dataset that has $27,386$ examples for the training set and $6,844$ examples for the test set, both with a 50-50\% ratio of positives and negatives, and with separate objects in the training and test sets. After training our instance model on this dataset, we evaluate the accuracy of the model on the test set. \tab{tab:accuracy} shows our model obtains an overall accuracy of $64.3\%$, which is significantly above chance ($50.0\%$).

As discussed in Section~\ref{sec:kway}, a compelling practical application of this approach is to recognize object instances by touch from a pool of potential candidate images. Such cases arise frequently in industrial and logistics settings, such as manufacturing, where a robot might need to recognize which out of a set of possible parts it is handling, or warehousing, where a robot must retrieve a particular object from a shelf.

We simulate this situation in a $K$-shot classification test by providing the model with tactile readings from a single grasp in the test set paired with $K$ different object images. One of those object images corresponds to the actual object the robot was grasping at the time and the other $K-1$ images are of objects randomly selected from the test set. For each pair we then evaluate how confident our model is that the tactile readings and object image correspond. We rank the objects by their confidences, and measure how many guesses it took our model to select the correct object. We perform this evaluation for both $K=5$ and $K=10$ objects. 

Prior work has suggested canonical correlation analysis (CCA) ~\cite{cca} as a method for cross-modal classification ~\cite{arandjelovic18} and we use this prior approach as a baseline for our method. Note, however, that the central hypothesis we are testing is whether cross-modal instance recognition of visual instances from touch is possible \emph{at all}, and this baseline is provided simply as a point of comparison, since no prior work tests this particular type of cross-modal recognition. In \fig{fig:roc5} and \fig{fig:roc10}, we show the accuracy of the model in both $K=5$ and $K=10$ settings. The accuracy is visualized as a function of the number of guesses, measuring how often the correct object is guessed within the first $N$ guesses.

\begin{figure*}[t]
  \centering
  \includegraphics[width=\linewidth]{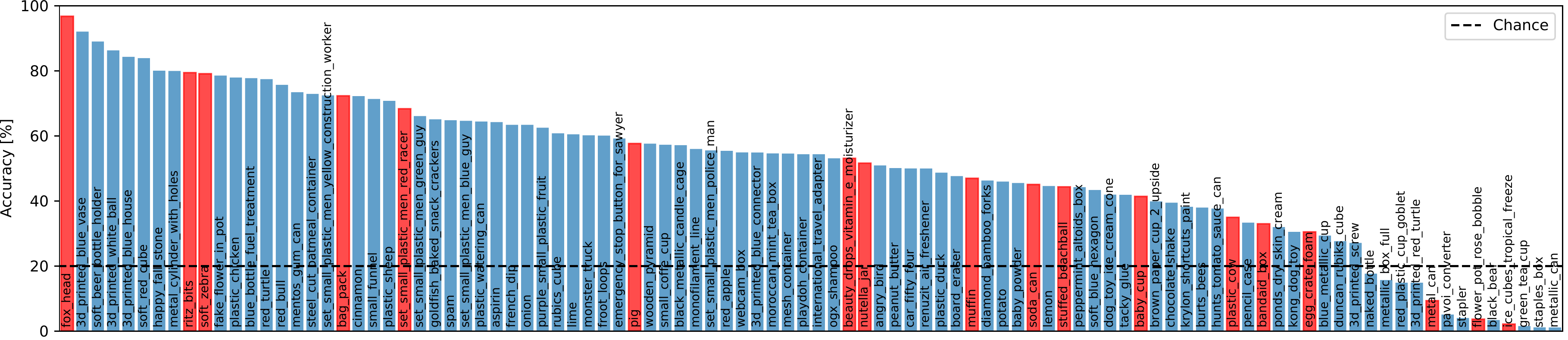}
  \caption{Prediction accuracy by object for the $K = 5$ first-shot classification when considering $(T, I)$ pairs from all possible objects. Red bars indicate test objects and the blue bars training objects. The red and blue bars are distributed fairly evenly, indicating that our model does not perform much worse on the unseen test objects compared to the training objects.}
    \label{fig:accall}
\end{figure*}

\begin{figure}[t]
  \centering
  \includegraphics[width=0.90\linewidth]{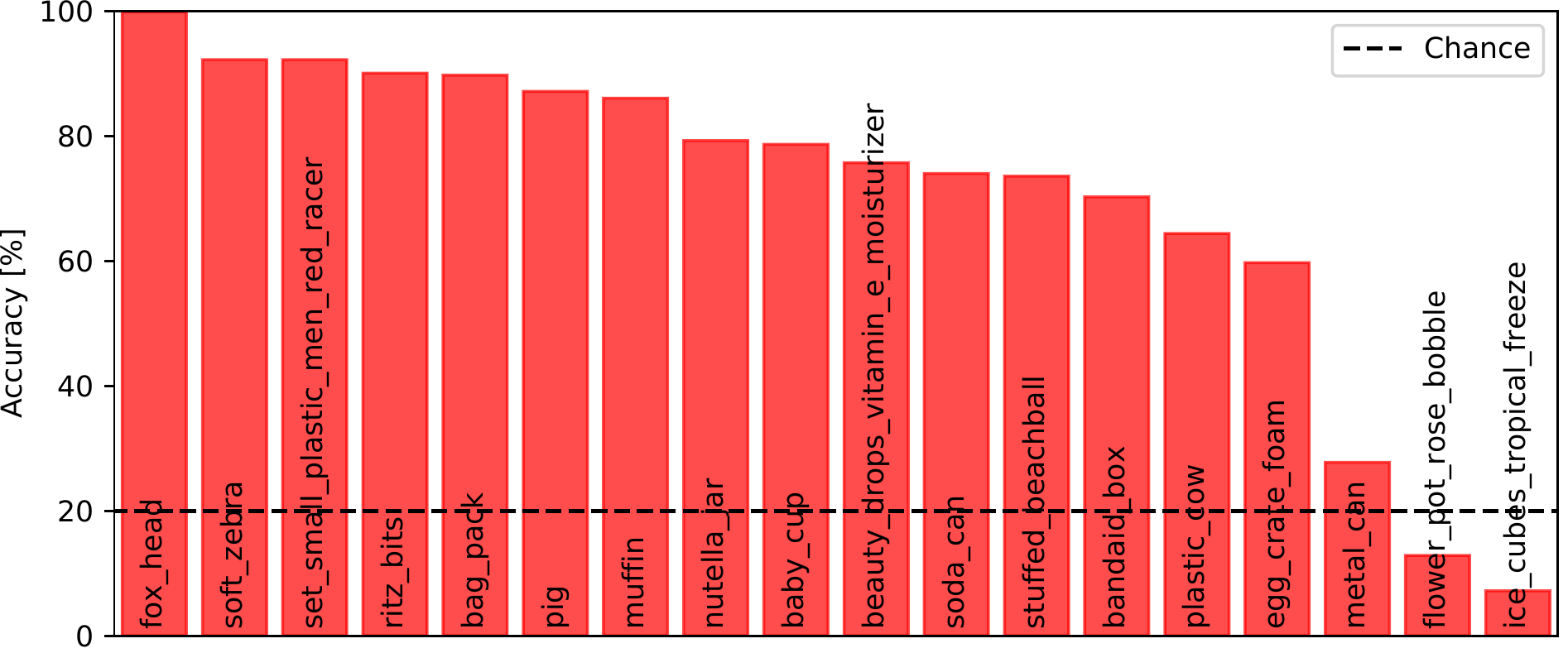}
  \caption{Prediction accuracy by object for the $K=5$ first-shot classification when considering $(T, I)$ pairs exclusively formed from objects in the test set. For comparisons involving entirely unseen objects, our model is still able to identify a majority of the objects with high accuracy.}
    \label{fig:acctest}
\end{figure}

We also look at our model's performance with respect to each object through a first-shot classification task. Similar to the 5-shot object identification task, we once again generate 5 pairs of tactile readings and object images, but if our model's first guess is incorrect we note what other object the true object was mistaken for, rather than continuing to guess. Accuracy when considering pairs generated from only objects in the test set is shown in \fig{fig:acctest} and accuracy when considering pairs generated from all possible objects in both the training and test set is plotted in \fig{fig:accall}. When looking at the accuracies for the case in which we consider all possible objects, we notice that the distribution of performances for objects in the training set does not seem to differ significantly from the distribution of performances for objects in the test set, which suggests that our model learns a generalizable approach for object identification.

\subsection{Comparison to Human Performance} 
Since providing a baseline for the performance of our model is difficult due to the lack of prior work in this problem setting, we also compare our model's performance to that of humans. Here, we evaluate the performance of undergraduates at the University of California, Berkeley on the exact same 5-shot classification task as above. Subjects are shown GelSight tactile readings taken by the robot and, after a training period where they are provided with example tactile-visual associations, are asked to predict which object corresponds to a particular tactile reading. We collect 420 trials from 11 volunteers and their performance relative to the other methods on the 5-shot classification task can be seen in \fig{fig:roc5}. Our model outperforms humans at this object identification task, although we should note that humans are not accustomed to observing objects in this manner, as we directly use our sense of touch rather than looking at the deformation of our fingers. 

We hypothesize this object identification task to be so difficult because a 2-dimensional image cannot fully capture the physical characteristics of an object. When grasping an object, it is possible for the object to be in a different orientation than what is shown by the image, and it is also possible for a given object to have drastically disparate tactile readings depending on where that object is being grasped. To do well on this task requires one to infer not only what material(s) an object is made of but also possible locations at which the object might be grasped, all based on the limited information provided by a 2-dimensional image.


\section{Discussion and Future Work} 
\label{sec:conclusion}

	In this work, we propose the cross-modality instance recognition problem formulation. This problem statement requires a robot to infer whether a given visual observation and tactile observation correspond to the same object. A solution to this problem allows a robot to recognize objects by touch: given pictures of candidate objects, the robot pairs the tactile readings with each and recognizes the object based on which image is assigned the highest probability of a match. We propose to address this problem by training a deep convolutional neural network model on data collected autonomously by a robotic manipulator. The aim of our experiments is to test whether it is possible to utilize tactile sensing to recognize object instances effectively. In our experiments, a robot repeatedly grasps each object, associating each of the recorded images with each of the tactile observations on that object creating positive examples. All pairs of observations across different objects are then labeled as negative examples. This procedure is largely automatic, since the robot can collect a large number of grasps on its own, which provides us with an inexpensive method for collecting training data.

Our experimental results demonstrate that it is indeed possible to recognize object instances from tactile readings: the detection rate of objects is substantially higher than chance even for novel objects and our model outperforms alternative methods.

There are a number of promising directions for future work. In this work, we consider only individual grasps but a more complete picture of an object can be obtained from multiple tactile interactions. Integrating a variable number of interactions for a single object recognition system is therefore a promising direction for future work. Furthermore, extending upon our proposed approach within a robotic manipulation framework is an exciting direction for future research: by enabling robots to recognize objects by touch, we can image robotic warehouses where robots retrieve objects from product images by feeling for them on shelves, robots in the home that can retrieve objects from hard-to-reach places, and perhaps a deeper understanding of object properties through multi-modal training.


\section*{Acknowledgements} 
We thank Andrew Owens for his insights about multi-modal networks and his suggestions for the manuscript. We also thank Wenzhen Yuan and Edward Adelson for providing the GelSight sensors.



\bibliographystyle{IEEEtran}
\bibliography{paper-tactile3}  

\end{document}